# A Scalable Machine Learning Approach for Inferring Probabilistic US-LI-RADS Categorization

Imon Banerjee, Ph.D[1], Hailey H. Choi, M.D[2], Terry Desser, M.D[2], Daniel L. Rubin, M.D[1,2]
[1]Department of Biomedical Data Science, Stanford University School of Medicine
Medical School Office Building, Stanford CA 94305-5479;
[2]Department of Radiology, Stanford University School of Medicine
Stanford CA 94305-5479

**ABSTRACT**

*We propose a scalable computerized approach for large-scale inference of Liver Imaging Reporting and Data System (LI-RADS) final assessment categories in narrative ultrasound (US) reports. Although our model was trained on reports created using a LI-RADS template, it was also able to infer LI-RADS scoring for unstructured reports that were created before the LI-RADS guidelines were established. No human-labelled data was required in any step of this study; for training, LI-RADS scores were automatically extracted from those reports that contained structured LI-RADS scores, and it translated the derived knowledge to reasoning on unstructured radiology reports. By providing automated LI-RADS categorization, our approach may enable standardizing screening recommendations and treatment planning of patients at risk for hepatocellular carcinoma, and it may facilitate AI-based healthcare research with US images by offering large scale text mining and data gathering opportunities from standard hospital clinical data repositories.*

**INTRODUCTION**

Hepatocellular carcinoma (HCC) is a worldwide public health problem, and ultrasound (US) examination of the liver is at present the only test universally recommended by liver societies for screening patients at risk for its development(1). The objective of an HCC screening and surveillance program is to identify HCC at an early, preclinical stage when it can potentially be cured either with local therapy or liver transplantation(2), and thus improve mortality. Despite the widespread use of US for evaluating the risk of HCC, until recently there have not been standardized guidelines for interpretation and reporting of liver findings. The clinical utility of HCC screening can be impaired if the US text reports are unclear, ambiguous, or variable from interpreter to interpreter. Recently (2017), the American College of Radiology (ACR) has developed the Ultrasound Liver Imaging Reporting and Data System (US LI-RADS) which contains standardized terms and management recommendations for sonographic examinations performed for HCC screening and surveillance(3). Given the millions of patients at high risk for HCC worldwide, one of the main goals of LI-RADS is to improve consensus among clinicians in categorization and management of findings on screening and surveillance ultrasound exams. In addition, standardized reporting can facilitate research, radiologist productivity, as well as quality assurance and improvement(4).

During the transition phase from non-LI-RADS reporting (unstructured free-text reports) to LI-RADS reporting, a difficult management question is: *how to follow a patient longitudinally (throughout multiple years) whose ultrasound screening exam was reported before LI-RADS was implemented*. Manual chart-review, although cumbersome, is one immediate solution, though it is costly and labor intensive. Alternatively, computerized inference of LI-RADS category directly from the report text may play a key role in standardizing treatment planning of HCC by automatically assigning LI-RADS categories to exams performed before the LI-RADS guideline was proposed. This would allow follow up of patients based on LI-RADS on a much larger scale. Moreover, in real time, it may also provide feedback, as US reports are dictated to help minimize inconsistencies in reporting of findings and recommendations for follow-up from one screening time point to the next. Inferring LI-RADS category for all the HCC screening and surveillance reports present in a large hospital data repository may facilitate AI-based healthcare research, with US images by offering large scale text mining and data gathering opportunities.

Several standard NLP-based machine learning methods (rule-based, dictionary-based, supervised learning) have been successfully applied to radiology reports in other clinical domains for extracting information or performing automatic report classification(5–9). Traditional NLP techniques have also been explored with clinical reports of screening studies for colorectal adenoma(10), inflammatory bowel disease(11), fatty liver disease(12), aortic aneurysm(6), and liver steatosis(13). However, a major limitation of these earlier studies is the requirement for large-scale human-labeled data or explicit hand-written linguistic rules that limit both the scalability and generalizability of the system. In contrast with the previous approaches, recent advances in NLP techniques can be leveraged for the automatic interpretation of free-text narratives by exploiting distributional semantics and can provide adequate generalizability by addressing linguistic variability(14,15). We propose a scalable NLP pipeline (**Figure 1**) for inferring LI-RADS coding by considering only the liver findings written in the narrative, unstructured form in US reports. The core novelties of the proposed pipeline are:

(1) In the US LI-RADS case-study, we translated our previously proposed algorithm(14,16) which was originally developed for parsing CT radiology reports. We applied the unsupervised algorithm that combines distributional semantics with the semantic dictionary mapping technique to create a dense vector representation of the liver findings written in narrative form;

(2) In addition to the previous studies, we exploited the word-vector embedding space to learn a LI-RADS vocabulary (semantic dictionary) using word similarity clustering (**Table 2**) from a large mixed corpus of recent LI-RADS coded reports and older non-LI-RADS reports, and mapped synonyms to lexicon terms for the non-LI-RADS reports to improve the scalability of the pipeline. The application of semantic dictionary mapping as the basis of non-LI-RADS report parsing helps to handle out-of-vocabulary words and increases the feature extraction efficiency for targeting the complex LI-RADS phenotyping task;

(3) Finally, the vector representations of liver findings along with the quantitative lesion measures extracted via regular expressions were utilized in an ensemble machine-learning setting to infer the three LI-RADS final assessment categories[1]: *LI-RADS1 - Negative*, *LI-RADS2 - Subthreshold*, and *LI-RADS3 – Positive* (3). An interesting feature of this study is that no human-label data were required for training the model, because the LI-RADS scoring was automatically extracted from the reports formatted with LI-RADS coding. Once the classifier was trained on the LI-RADS formatted reports, it was validated on both LI-RADS and non-LI-RADS reports (see **Results**).

**METHODS**

**Dataset**

With the approval of Institutional Review Board (IRB), we collected all the free-text radiology reports of abdominal ultrasound examinations done within our institution between August 2007 to December 2017. In total, there were 92,848 US reports collected over 10 years, and on average about 9,000 US exams were performed per year. Among them, 13,860 exams contained the HCC screening procedure code (Table 1). Prior to LI-RADS implementation, the impression section of the reports was used to summarize the core observations of the US scans as diagnosis/conclusion. Often a summary code-based system [1-9] where 1 means no risk and 9 means highest risk, was used to capture the overall status of the abdominal organs. However, there was no formal or standardized way to code the risk of hepatocellular carcinoma.

Starting from January 2017, the LI-RADS system was adopted in our institution for reporting examinations of the liver in patients at risk for HCC. In the year 2017, a total 1,744 abdominal US studies were coded with the LI-RADS reporting format where a unique LIRADS score was reported in the impression section. In Table 1, we present quarterly statistics of HCC screening and surveillance exams performed in the year 2017, according to assigned LI-RADS categories. However, there were 962 HCC screening studies (36%) that did not follow the LI-RADS template.

Among 1,744 US reports coded with LI-RADS template, we randomly selected 10% of the reports as our *validation dataset* (147 reports) and remaining reports (1277 reports) were used to *train* our machine-learning model. For training and validation, LI-RADS scores assigned by radiologists were available for all 1,744 reports. Yet, a major purpose of the study was to infer the LI-RADS score for 11,154 US exams done between 2007 – 2016. However, ground truth labels were missing for these reports since LI-RADS coding was not implemented at the time of reporting. Thus, for generating the *test set*, we asked two radiologists (specialized on reading US exams) to assign independently LI-RADS labels on a set of 216 reports by analyzing only the findings section of the reports. The reports were only included in the *test set* if the ensembled classifier model's predicted highest probability is either <0.5 (152 reports) or >0.9 (64 reports). The criteria for *test set* selection was mainly to capture the top and bottom level view of the model's performance.

**Table 1.** Statistic of our cohort: HCC screening studies between [2007 - 2017].

| Year 2007 - 2016 (without LI-RADS template) |
|---|
| 11,154 |
| **Year 2017 (without LI-RADS template)** |
| 962 |

---

[1] www.acr.org/Clinical-Resources/Reporting-and-Data-Systems/LI-RADS/Ultrasound-LI-RADS-v2017

| | Year 2017 (with LI-RADS template) | | | | |
|---|---|---|---|---|---|
| | Quarter 1 (Jan - March) | Quarter 2 (April -June) | Quarter 3 (July - Sept.) | Quarter 4 (Oct. - Dec.) | Total |
| LI-RADS 1 | 112 | 328 | 640 | 519 | 1589 |
| LI-RADS 2 | 3 | 21 | 37 | 32 | 93 |
| LI-RADS 3 | 2 | 4 | 32 | 30 | 62 |

**System methodology**

We developed a completely automatic system for inferring LIRADS categorization by looking only at the findings section of the US reports, which describe liver observations. The main modules of the system are presented in Figure 1 and the following subsections describe the working principle of the three core steps of our pipeline (highlighted in dotted line).

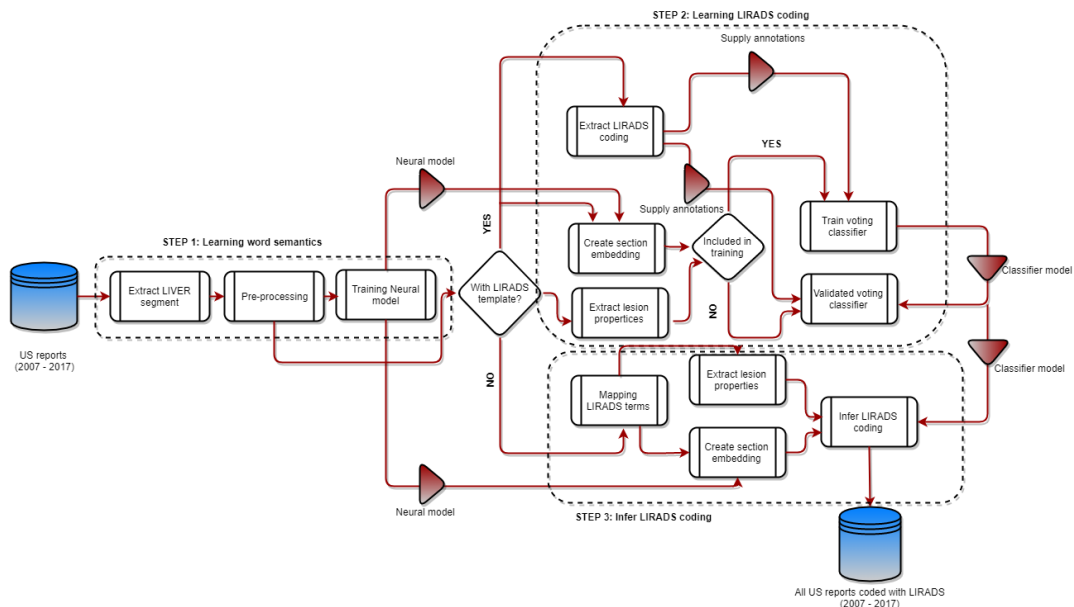

**Figure 1.** Proposed pipeline of LI-RADS inference system

**Step 1: Learning word semantics**

We found great variability in the linguistic style, reporting structure, and terms within our US report corpus collected over 2007-2017 within the same institution. Even with LI-RADS template, we saw term variations (use of synonyms) for reporting liver lesion features. For instance, in the LI-RADS reporting format, the same lesion was described by one reader as "*hyperechoic and thin septation*" while another rater describes it as "*hyperechogenic and multicystic*". The term variations expanded rapidly when considering US reports formatted without LI-RADS guidelines. In Step 1 of our method, the goal is to apply distributional semantics to learn the similarity between various terms used by the various radiologists for reporting the liver US findings across a large time window in an unsupervised way, with the intent of successfully mapping the terms that have been used before and after LI-RADS template implementation.

We developed a python-based LIVER section segmentation module that uses a combination of regular expressions to find and extract the liver findings from the whole report while maintaining dependencies between anatomical entities (e.g. all the lesions found within the liver would be extracted even if they are not describe in the same paragraph). In order to perform a valid experiment, we excluded the *Impression* section of the reports since the final LI-RADS assessment category is often reported explicitly in the *Impression*. The *Findings* section includes only the imaging characteristics of the liver abnormalities; thus, it does not contain any clear-cut definition of the LI-RADS final assessment classes. We extracted the liver section from all the abdominal US reports, including the non-HCC studies (on total 92,848 US reports); this enabled learning of more targeted similarity of words that can only be applied in the liver context.

All the extracted liver sections were transformed through a series of pre-processing steps to truncate the texts and to focus only on the significant concepts. This was accomplished by normalizing the texts to lowercase letters, word stemming, number to string conversion, and removing words of following types: general stop words, words with very low frequency [50], unwanted terms and phrases. Words appearing in either all reports or in very few reports have little or no value in document classification. We used the NLTK library(17) for determining a stop-word list and discarded them during indexing. Examples of the stop-words are: a, an, are,...,be, by,...,has, he,...,etc. In order to preserve the local dependencies, bigram collocations of all possible word-pairs were calculated for the entire pre-processed corpus based on Pointwise Mutual Information(18). The bigrams with fewer than 50 occurrences were discarded and the top 1000 bigram collocations were concatenated as a single word entities were to improve the accuracy of the word embeddings. A few examples of derived resultant bigrams from liver US are - '*ill_defined*', '*viral_hepatitis*', '*contrast_agent*', '*long_axis*', '*echo_texture*', '*capsul_smooth*'. We also used a dictionary-style controlled-term mapping technique, where publicly available CLEVER terminology[2] was used to replace common analogies/synonyms to create more semantically structured texts. We focused on the terms that described family, progress, risk, negation, and punctuations, and normalized them using the formal terms derived from the terminology. For instance, {'mother', 'brother', 'wife' .. } → '*FAMILY*', {'no', 'absent', 'adequate to rule her out' .. } → '*NEGEX*', {'suspicion', 'probable', 'possible' } → '*RISK*', {'increase', 'invasive', 'diffuse', .. } → '*QUAL*'.

After the pre-processing step, the corpus of 92,848 abdominal US reports was used to create vector embeddings for words in a completely unsupervised manner using the word2vec model. The word2vec adopts distributional semantics to learn dense vector representations of all words in the pre-processed corpus by analyzing the context. In other words, the vectors produced represent each word or phrase as a mathematical combination of the words and phrases surrounding it within a linear context window. We first constructed a vocabulary from our pre-processed tokenized corpus that contains 2000 unique words. Vector representations of words in the vocabulary is learned by a CBOW word2vec model using the Gensim 2.1.0 library(19). The CBOW word2vec model predicts a word given a context where the window size defines context. The loss function of CBOW is: $E = -v_{w_o}'.h + \log \sum_{j=1}^{V} \exp(v_{w_j}'.h)$, where $w_o$ is the output word, $v_{w_o}'$ is its output vector, $h$ is the average of vectors of the context words, and V is the entire vocabulary. Once the model constructed the vectors, we used the cosine distance of vectors to denote similarity, thereby deriving analogies. As the training algorithm, we used Negative Sampling where the cost function is: $E = -\log \sigma(v_{w_o}'.h) - \sum_{w_j \in \omega_{neg}} \log \sigma(-v_{w_j}'.h)$, where $\omega_{neg}$ is the set of negative samples, $w_o$ is the output word, $v_{w_o}'$ is its output vector and h is the average of vectors of the context words.

To test the semantic validity of the trained model, we derived the synonyms of the terms belonging to the LI-RADS lexicon set by computing the similarity score between the word vectors as cosine similarity which is inner product on the normalized space that measures the cosine of the angle between two words: $Similarity = \frac{A \cdot B}{\|A\|\|B\|} = \frac{\sum_{i=1}^{n} A_i B_i}{\sqrt{\sum_{i=1}^{n} A_i^2} \sqrt{\sum_{i=1}^{n} B_i^2}}$. In Table 2, we only report the synonyms derived by the model with $> 0.80$ similarity score which shows that the model can capture the basic terms variation for liver reporting.

**Table 2.** Synonyms of LI-RADS terminology derived by the model.

| Categories | LI-RADS Lexicon | Synonyms generated |
|---|---|---|
| Echogenicity | hyperechoic | hyperechogenic, hyperecho |
| | isoechoic | isoecho |
| | hypoechoic | hypoechogenicity, hypoechogen, hypoecho |
| | cystic | anecho, anechoic |
| | nonshadowing | non_shadowing |

---

[2] https://github.com/stamang/CLEVER/blob/master/res/dicts/base/clever_base_terminology.txt

| Doppler vascularity | hypovascular | nonenhancing |
| --- | --- | --- |
| | avascular | nonvascular |
| | hypervascular | hypervascularity |
| Architecture | septation | septat, septations, multicystic, septa, complex_cyst, intern_septation, thin_septation, multispet, reticul, fishnet, multilocul |
| | complex | complicated, solid_and_cystic |
| Morphology | lobulated | bilobe, macrolobulated, microlobulated |
| | round | oval, rounded, ovoid, oblong |
| | ill-define | vague, indistinct |
| | exophytic | bulge |
| | well_defined | well_circumcribed, marginated |

**Step 2: Learning LI-RADS coding**

After learning the semantics of individual words that are used for characterization of liver findings, the next challenge was *how to train the machine to infer LI-RADS coding by exploiting the existing knowledge*. Our goal was to minimize the human annotator involvement as much as possible. Nevertheless, we needed an annotated dataset to teach the machine the LI-RADS assessment principles. Therefore, we used a simple regular expression to extract the recorded LI-RADS categories from the impression section of those reports that used the LI-RADS template. The success rate of the LI-RADS category extraction was 100% since a very standard reporting format is used which is trivial to code using regular expressions. This helped generate a training set size of 1744 annotated reports.

Finally, the liver section embedding was created by averaging the word vectors generated via the trained model. According to previous studies(20), averaging the embeddings of words in a sentence/paragraph has been proven a successful way of obtaining embeddings, where each section vector was computed as: $v_{doc} = \frac{1}{\|V_{doc}\|} \sum_{w \in V_{doc}} v_w$, where $V_{doc}$ is the set of words in the report and $v_w$ refers to the word vector of word $w$. The dense and relatively lower dimensional (compare to bag-of-words) numerical representation of the texts reflecting liver characterizations were used as feature vector to train efficiently a supervised classifier that can leverage the LI-RADS labels extracted earlier to learn the coding principle.

An interesting LI-RADS guideline for differentiating LI-RADS 2 (Subthreshold) from LI-RADS 3(Positive) is that the lesion must be ≥10mm to qualify as LI-RADS 3. Therefore, there is a clear need to capture the quantitative properties of lesion with maximum dimension. Theoretically, the word2vec model should also learn to interpret the difference between numerical values. However, due to the limitation of the training corpus size (usually in generic domain trained with millions of text snippets), the model was not able to learn every numerical combination, instead learning them all as generic number set. For example, the top similar word for '*one*' was '*nine*', and the set of top nine similar words contained all one-digit numbers. Thus, we extracted another set of features from the liver section using linguistic rules that captured two quantitative measurements: (1) number of lesions present in the liver, and (2) long-axis length of the largest lesion. For long-axis length, we extracted all 3D, 2D and 1D lesion measures mentioned in the text, normalized unit to mm, and identified the maximum value.

Finally, we trained an ensemble classifier - a meta-classifier targeted to predict LI-RADS scoring by combining two different classifiers based on weighted voting: (1) *section embedding classifier*: takes vector representation of the liver section as input, and (2) *lesion measure classifier*: takes the two quantitative lesion measures as input. Logistic regression with stochastic average gradient solver was used to build the section embedding classifier while decision tree classifier was applied on the quantitative lesion measures. We believe that the ensembles of these classifiers can be more powerful than any of the individual members since nature of both classifiers is complimentary in the sense that the possible errors made by the classifiers should be uncorrelated. For instance, if the section embedding classifier failed to differentiate LI-RADS 2 and LI-RADS 3, the lesion measure classifier should be able to distinguish them. To manage the class imbalance issue within the *training dataset* (see **Dataset** section), we performed over-sampling using the Synthetic Minority Over-sampling Technique (SMOTE) for the under-represented LI-RADS class 2 and 3.

**Step 3: Infer LI-RADS coding**

In this study, one of our final objectives were to train the ensemble classifier with the features extracted only from the US reports coded with LI-RADS template (*training set*) and use the model to infer the LI-RADS coding for all the remaining reports that were not written following the LI-RADS guideline (*test set*). We extracted the same set of features from the testing set by using the trained word2vec model and the lesion measure extraction rules. We found that the extraction rules are easily generalizable for different report formats, since lesion measures are always documented using somewhat similar styling (e.g. 3D - 1.2 × 1.3 × 0.9cm, 2D - 1.2 × 1.3cm, 1D - 1.2cm).

However, before generating the section embedding, we replaced all the synonyms of LI-RADS lexicon terms with the root term (see Table 2) to create normalized section embeddings for the test set. In the current setting, we only used 1,589 US reports coded with LI-RADS to infer the LI-RADS scoring for 11,154 US exam reports done between 2007 -2016. In the following section, we report the performance of the model on both *validation (147 reports with LI-RADS template)* and *test dataset (216 reports without LI-RADS template).*

**RESULTS**

**Performance with LI-RADS template: validation of the model**

In clinical practice, inter-rater disagreement is often reported regarding the final assessment of LI-RADS categories. We compared head-to-head the performance of two independent radiologists (rater 1 and rater 2) specialized in reading US exams against our proposed machine learning model. We considered the LI-RADS category reported by the original interpreting radiologist as the true label. The performance of human raters and the proposed model on the *validation set* (147 reports) is reported in Table 3 as *precision*, *recall*, and *f1 score*: harmonic mean of precision and recall value. In this experiment, there was a high interrater agreement for LI-RADS scoring as the raters are from the same institution and mutually agreed upon a set of internal rules beforehand. We also present performance of a machine learning classifiers only using unigrams as features which can be considered as the baseline machine learning performance to be compared with the proposed ensemble classifier.

While the reported performance of the baseline with unigrams (measured as f1 score) is 0.52 averaged over all three LI-RADS classes, the proposed model f1 score is 0.90 which is comparable with the human rater's performance in annotating the same validation set. This demonstrates that our model was able to learn the significant facets of the liver findings for characterizing LI-RADS. The precision of the proposed model's categorization is equal to the highest scoring human rater while the recall is slightly low. It is important to analyze the LI-RADS class-wise prediction to understand the scoring trend of the machine as well as of the raters. However, because of averaging over all the LI-RADS classes, these observations can only represent an overall accuracy, but cannot provide a proper reflection on the scoring.

**Table 3:** Comparison of machine-generated annotations with two human raters: reports with LI-RADS template

|  | Machine inferred annotation | | Human annotation | |
|---|---|---|---|---|
|  | BOW classifier | Proposed model | Rater 1 | Rater 2 |
| **Average precision** | 0.59 | 0.93 | 0.92 | 0.93 |
| **Average recall** | 0.49 | 0.88 | 0.92 | 0.92 |
| **Average f1 score** | 0.52 | 0.90 | 0.92 | 0.92 |

In Figures 2 and 3, we present the class-wise prediction performance of the human raters and machine learning classifiers as confusion matrix where x-axis represents predicted outcome and y-axis represents the true labels, and numerical values show class-wise sample in each cell. In order to show the proficiency of final ensemble model, we also present the individual confusion matrix for lesion measure classifier (Figure 3.a) and section embedding classifier (Figure 3.b). Lesion measure classifier performed better in classifying LI-RADS 3 while section embedding classifier outperformed in LI-RADS 1 and 2 categorization. As expected, the ensemble classifier combined the two individual classifiers' strengths and achieved uniform classification power for all three LI-RADS classes, which is comparable to the human raters. An exciting outcome is that the proposed model outperformed human raters in classifying LI-RADS 2 and LI-RADS 3 categories.

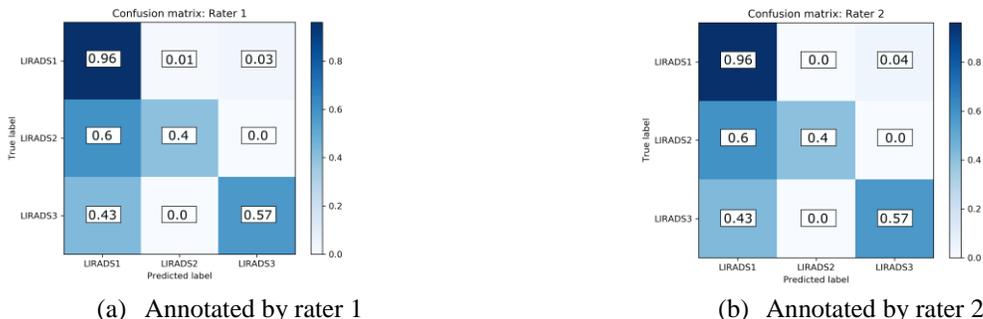

(a) Annotated by rater 1     (b) Annotated by rater 2

**Figure 2.** Human raters' performance for inferring LI-RADS coding looking at the LIVER findings (147 cases).

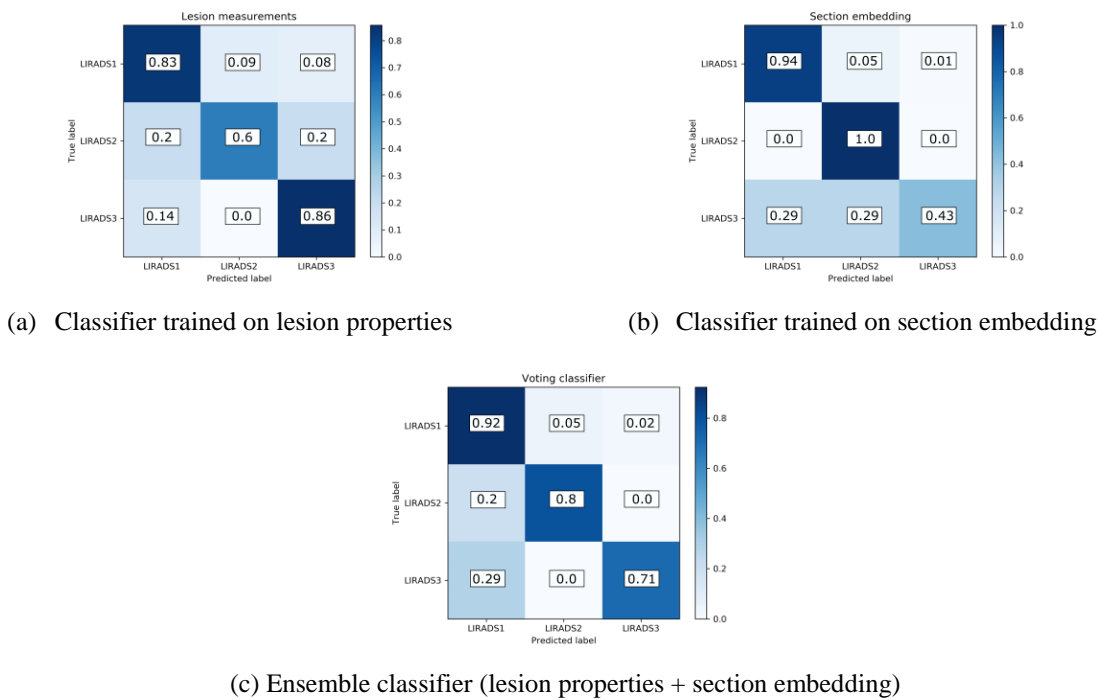

(a) Classifier trained on lesion properties     (b) Classifier trained on section embedding

(c) Ensemble classifier (lesion properties + section embedding)

**Figure 3.** Machine performance for inferring LI-RADS coding looking at the LIVER findings (147 cases).

In Table 4, we present the reasons for disagreement between the human raters and the original image reader for a few cases, and also highlight the probabilistic prediction value of the proposed model. This shows that there is no well-defined ground truth for labeling, and individual experts may have different opinions. The model prediction is mostly consistent with the scoring of the original reader (with one exception); however, the class probability value clearly shows that model is deciding between reader's defined label and raters' observation. Being trained on the dataset which involves multiple US image readers with significant disagreement for some cases, the probabilistic prediction of the model represents an interesting integrated view of multiple individuals for these complicated cases.

**Table 4:** Examples of disagreement between original US image reader and the raters. Also includes the probabilistic and discrete prediction outcome of the proposed algorithm for those cases.

| Liver Segment | Original coding | Machine derived probability | Imputed label | Rater 1 | Rater 2 | Reason |
| --- | --- | --- | --- | --- | --- | --- |

| | | 1 | 2 | 3 | | | | |
|---|---|---|---|---|---|---|---|---|
| liver length: 14.2 cm. liver appearance: mild steatosis. segment 5 lesion, previously characterized as a hemangioma on february 2014 mr now measures 1.0 x 1.0 x 0.9 cm, previously 0.7 x 0.6. previously seen optn class 5a lesion on february 2014 mr is not well seen on ultrasound. null a small right hepatic lobe cyst measures 7 x 6 x 7 mm, previously 10 x 9 x 9 mm. no new hepatic lesions. | 3 | 0.56 | 0.06 | 0.38 | 1 | 1 | 1 | Previously characterized as hemangioma, therefore should be categorized as benign |
| liver length: 17.6 cm. liver appearance: normal. liver observations: 0.6 x 1.1 x 1.3 cm hyperechoic focus in the right liver was minimal flow likely representing a hemangioma or focal fat. liver doppler: hepatic veins: patent with normal triphasic waveforms. | 1 | 0.55 | 0.06 | 0.4 | 1 | 2 | 3 | Includes hemangioma or fat (both benign), but this is not definite and needs characterization |
| liver length: 12.1 cm. liver appearance: mild steatosis. hypoechoic left hepatic lobe lesion measures 1.2 x 0.5 x 0.7 cm, decreased from 3/8/2017 ct and not significantly changed from more recent pet/ct. | 1 | 0.45 | 0.10 | 0.44 | 1 | 3 | 3 | Observation is stable from prior imaging, but not definitely benign |
| liver length: 16 cm. liver appearance: severe steatosis. null no surface nodularity. liver observations: 1.7 x 1.4 x 1.0 cm hypoechoic focus in the gallbladder fossa likely reflects focal fatty sparing. null no surface nodularity. liver doppler: hepatic veins: patent with normal triphasic waveforms. | 3 | 0.43 | 0.17 | 0.40 | 3 | 1 | 1 | Fatty sparing is benign |

**Performance without LI-RADS template: test on reports with different linguistic style**

Finally, we used the trained ensemble model to infer the LI-RADS score for 11,154 US exams done between 2007 - 2016. To verify the model's performance, we used the **Test set** created by two independent radiologists who assigned LI-RADS labels on a subset of 216 reports coded without LI-RADS template where the model's predicted highest probability is either <0.5 (152 reports) or >0.9 (64 reports) (see **Dataset** section).

For the labels provided by rater 1, the proposed model scored an average of 0.77 precision and 0.63 recall, while for rater 2, the average precision was 0.74 and recall was 0.62. Figure 4 represents class-wise performance for individual raters as confusion matrix. The figure shows that the model achieved >85% accuracy in classifying LI-RADS 2, 60% in LI-RADS 3, and 57% for LI-RADS 1 for both raters. Actually, the model is overestimating LI-RADS 1 as LI-RADS 2, and underestimating LI-RADS 3 as LI-RADS 2. This is probably due to the inconsistency in the training dataset and accumulating confusion between different readers while classifying adjacent LI-RADS classes. As a matter of fact, there was also significant disagreement between the original image readers and the raters upon LI-RADS 2 and 3 categorization for the report that followed the LI-RADS template as seen in Figure 2.

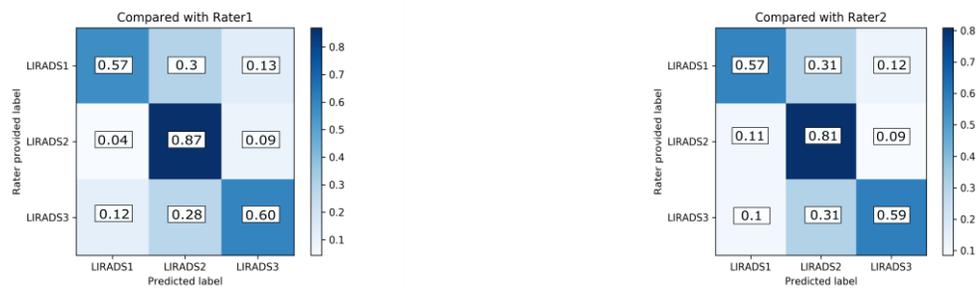

**Figure 4:** Performance of the proposed model for inferring LI-RADS coding: from 216 reports formatted without LI-RADS template.

## DISCUSSION

In this study, we proposed an automatic pipeline to infer the LI-RADS category of the free-text liver ultrasound reports. The model was trained on the reports created using a LI-RADS template and validated on both LI-RADS (147 reports) and non-LI-RADS (216 reports) coded reports. We translated our previously proposed pipeline(14,16) to generate the vector representation of free-text liver finding documented in the US reports, and we designed an ensemble machine-learning model that considered the section vectors generated via Step 1 - 2 and quantitative lesion measures (Step 2) as inputs and inferred the LI-RADS score.

The core methodological challenge and contribution of our work was how to transfer the machine learning model that has been trained on the LI-RADS reports, and to apply on a different set of US report corpus that were written before the establishment of LI-RADS standard. This is because, in the LI-RADS reports, the radiologists supposedly used only a pre-defined set of terms for describing liver and lesion characteristics following the LI-RADS guidelines, whereas the non-LI-RADS reports were collected over a wide-span of time (August 2007 to December 2016) which may not only have a wide diversity in the reporting format, but also contains huge variability in the linguistic style, and reporting terms. Thanks to the use of distributional semantics approach, our model was able to learn the semantic similarity between various terms used by multiple radiologists for reporting the liver US findings across a large time window in an unsupervised way, and successfully mapped the terms that had been used before and after LI-RADS template implementation for creating the vector embedding of the reports. In order to capture the quantitative lesion measures, we also created a set of regular expressions that can extract the number of lesions present in the liver and long-axis length of the largest lesion that were generalizable to the non-LI-RADS reports.

The experimental results demonstrated that, on the LI-RADS coded reports, our ensemble classifier model (section embedding + lesion measures) achieved comparable performance with the highest scoring human raters. Due to the absent of ground truth for the non-LI-RADS reports, we compared the model against the labels provided by two raters, and the model scored decently when being validated against both: rater 1: 0.77 precision and 0.63 recall; rater 2: 0.74 precision and recall was 0.62. Reduced performance on the non-LI-RADS test set can be influenced by two main factors. First, for capturing a comprehensive view, we selected a mixture of the most difficult and most trivial test cases for the experiment, where the number of difficult cases were larger than trivial cases (>2x). Second, there were no ground truth labels available for the test set. Therefore, the model was only tested against two individual raters while a moderate amount of disagreement has been observed between original images readers and the same raters, even for classifying LI-RADS coded reports (see **Figure 2**). Thus, there were no absolute performance validation for non-LI-RADS reports, and only a comparative assessment was possible.

Our current study is a limited single-center, retrospective study; however, our results suggest that our model can be generalizable for inferring LI-RADS scores in a larger scale. Thus, in the transition from non-LI-RADS reporting to LI-RADS reporting for HCC screening and surveillance ultrasounds, our method may help in longitudinal follow-up of patient cohorts, whose ultrasound screening and surveillance exams were performed prior to LI-RADS implementation. In the prospective future studies, we will work on combining multiple raters' views for labeling the same reports for training the model more efficiently. We will also evaluate the stability of the liver lesion by tracking it through longitudinal ultrasound reports.

## ACKNOWLEDGEMENTS

This work was supported in part by grants from the National Cancer Institute, National Institutes of Health, 1U01CA190214 and 1U01CA187947.